\documentclass[10pt,twocolumn,letterpaper]{article}

\usepackage{cvpr}              
\usepackage{graphicx}
\usepackage{amsmath}
\usepackage{amssymb}
\usepackage{booktabs}
\usepackage{multirow}
\usepackage[table]{xcolor}
\usepackage{subcaption}

 \usepackage[T1]{fontenc}
\usepackage{newtxtext,newtxmath}
\usepackage[pagebackref,breaklinks,colorlinks,allcolors=cvprblue]{hyperref}

\usepackage{multirow}

\definecolor{cvprblue}{rgb}{0.21,0.49,0.74}
\usepackage[pagebackref,breaklinks,colorlinks,allcolors=cvprblue]{hyperref}

\def\paperID{*****} 
\def\confName{CVPR}
\def\confYear{2026}

\title{Can We Build Scene Graphs, Not Classify Them? FlowSG: Progressive
Image-Conditioned Scene Graph Generation with Flow Matching}

\author{Xin Hu$^{1,2}$, Ke Qin$^1$, Wen Yin$^1$, Yuan-Fang Li$^3$, Ming Li$^4$, Tao He$^1$\thanks{Corresponding author.}\\
$^1$The Laboratory of Intelligent Collaborative Computing of UESTC\\
$^2$Tianfu Jiangxi Laboratory, $^3$Monash University\\
$^4$Guangdong Laboratory of Artificial Intelligence and Digital Economy (SZ) \\
{\tt\small \{xh1m22,qinke,yinwen1999\}@uestc.edu.cn} \\
\tt\small {yuanfang.li@monash.edu,ming.li@u.nus.edu,tao.he01@hotmail.com}
}

\begin{document}
\maketitle
\begin{abstract}
 Scene Graph Generation (SGG) unifies object localization and visual relationship reasoning by predicting boxes and subject–predicate–object triples. Yet most pipelines treat SGG as a one-shot, deterministic classification instead of a genuine {progressive, generative} task. We propose \textbf{FlowSG}, which recasts SGG as continuous-time transport on a {hybrid} discrete–continuous state: starting from a {noised graph}, the model progressively grows an image-conditioned scene graph through constraint-aware refinements that jointly synthesize nodes (objects) and edges (predicates). Specifically, we first leverage a VQ-VAE to quantize a {scene graph} (e.g., the continuous visual features) into compact, predictable tokens; a graph Transformer then (i) predicts a conditional velocity field to transport continuous geometry (boxes) and (ii) updates discrete posteriors for {categorical tokens (object features and predicate labels)}, coupling semantics and geometry via flow-conditioned message aggregation. 
Training combines flow-matching losses for geometry with a discrete-flow objective for tokens, yielding few-step inference and plug-and-play compatibility with standard detectors/segmenters. Extensive experiments on VG and PSG under closed- and open-vocabulary protocols show consistent gains in predicate R/mR and graph-level metrics, validating the mixed discrete–continuous generative formulation over one-shot classification baselines, e.g.,  an average improvement of about {$3$}  points over the SOTA USG-Par.
\end{abstract}    
\section{Introduction}
\label{sec:intro}

Scene Graph Generation (SGG) \cite{cong2023reltr,tang2019learning} parses an image into a structured graph in which nodes are objects and directed edges encode their relations. This representation closes the gap between low-level perception and high-level reasoning, supporting applications such as visual question answering \cite{zakari2025vqa}, image retrieval \cite{song2017deep}, cross-modal learning \cite{Liu_2025_CVPR, Yan_2025_CVPR}, and emotion reasoning \cite{yin2025knowledge,yin2025tical}. The subsequent Panoptic SGG (PSG) \cite{yang2022panoptic, hu2025spade} further extends SGG by incorporating pixel segmentations of both countable objects and amorphous stuff, while simultaneously predicting relationships.

\begin{figure}[t]
\centering
\includegraphics[width=0.45\textwidth]{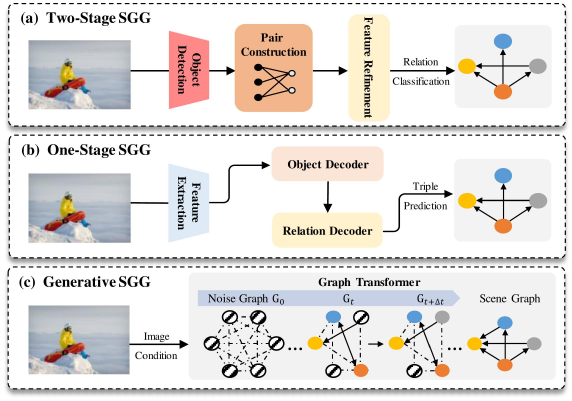}
\caption{Comparison of recent SGG paradigms. (a) \textbf{Two-stage}: a pre-trained detector proposes objects and enumerates human–object pairs; a relation head refines multi-stream features to classify predicates. (b) \textbf{One-stage}: objects and predicates are detected jointly in a single pass, followed by a matching step to attach predicates to object pairs. (c) \textbf{Ours (generative)}: given an image and an initially noisy graph, an image-conditioned denoiser iteratively refines the graph to sample a coherent scene graph.
}
\label{fig:intro}
\end{figure}

Broadly speaking, most SGG/PSG models follow a \emph{one-shot} formulation: either (i) a two-stage pipeline (see Fig. \ref{fig:intro}(a))~\cite{zhou2024openpsg,min2025vision} that first detects objects and then classifies relations over pre-defined pairs using GNN~\cite{xu2017scene} or transformers~\cite{fu2025hybrid,he2026lifelong}, or (ii) a one-stage set prediction (see Fig. \ref{fig:intro}(b)) that emits all subject–predicate–object triplets in a single forward pass~\cite{yang2022panoptic,fu2025hybrid}. Despite architectural differences, both families make \emph{deterministic} pair-selection and graph construction decisions within a single pass, eschewing iterative, progressive graph generation.


In effect, these methods perform scene graph \emph{prediction} rather than \emph{generation}: visual features are mapped directly to a final graph without an explicit, generative process. We deem that this paradigm has three drawbacks. \emph{First}, although internal feature refinement (e.g., via GNNs or Transformers) may occur, the decisive relation assignments are made in a single forward pass, leaving no mechanism for iterative correction, i.e., mis-paired entities or misclassified predicates cannot be revisited in light of emerging graph-level evidence.  \textit{Second}, semantic and geometric cues are treated as static inputs rather than jointly evolving states, preventing mutual refinement across iterations. \textit{Third}, enforcing graph-level constraints {(e.g., spatial transitivity \cite{spatialvlm})} is difficult when relations are scored independently, resulting in globally inconsistent scene graphs. We posit that, absent a genuinely generative, progressive construction of scene graphs, these methods are intrinsically limited in producing coherent, globally consistent SGs. This insight raises a research question: \emph{Can scene graph generation be reformulated as a truly progressive, generative problem rather than a one-shot classification task?}

With this question in mind, we advocate a \emph{generative} perspective in which scene graphs are \textit{constructed} through progressive graph refinement. Recent advances in  Flow Matching \cite{gat2024discrete,vignac2023digress} for discrete graph generation demonstrate that iterative denoising of node/edge states yields high-fidelity, constraint-aware graphs, such as molecule graphs generation \cite{hassan2024flow,eijkelboom2024variational}. In this work, we cast SGG as continuous-time transport on a hybrid state space by flow matching to evolve an noised graph into a coherent, image-conditioned scene graph via a sequence of constraint-aware refinements that jointly generate scene graph.

Concretely, we propose a  generative \textbf{FlowSG} framework,  which models a scene as a hybrid graph coupling symbolic semantics with geometry: each node stores a discrete object label and continuous box parameters, and each edge carries a discrete predicate. To curb the cost of high-dimensional visual features, we first compress object apperance features into compact codes via vector quantization \cite{van2017neural}. For generation, we adopt flow matching: a time-conditioned vector field drives a probability-flow ODE \cite{qin2024defog} that deterministically transports corrupted geometry and semantics toward the data manifold in a few large steps than conventional diffusion process {\cite{gat2024discrete,ho2020denoising}}. 
In parallel, discrete tokens are updated by predicting masked categorical posteriors under a discrete-flow objective \cite{gat2024discrete}, while the continuous geometry boxes are learned via conditional flow matching on Gaussian-perturbed boxes \cite{eijkelboom2024variational}.  In practice, the network directly regresses the conditional flow for boxes and the clean categorical distributions for tokens, fusing them at each step to yield fast, few-step synthesis with strong semantic–geometric consistency.

Moreover, scene graphs are typically sparse and exhibit heavy-tailed degree distributions, making vanilla message passing under-expressive and degree-sensitive. 
We first leverage {relation-modulated attention} in a FiLM~\cite{vignac2023digress,perez2018film} manner to inject predicate semantics into the attention computation, selectively amplifying relation-consistent neighbors while suppressing spurious connection. 
We then introduce a {Flow-conditioned Message Aggregation} (FMA) block: FMA computes a bank of neighborhood moments (e.g., degree, variance, skewness) and applies learned degree-aware scalers to regularize messages across low-/high-degree nodes, yielding richer and more stable updates.
Conditioned on frozen image features, the graph transformer iteratively denoises node geometry and edge predicates, producing clean posteriors and geometry flows that remain consistent with the visual evidence.

In summary, our contribution can  be fourfold as below:
\begin{itemize}
  \item \textbf{New Paradigm.} We reformulate SGG as progressive generation in a mixed space, jointly denoising discrete semantics and continuous geometry to enable global, structure-constrained refinement.
  \item \textbf{Hybrid Flow Matching.} We learn a time-aware velocity field that transports boxes via \emph{conditional flow matching} and updates discrete tokens via a \emph{discrete-flow} objective, where the two streams are tightly coupled at each  step.
  \item \textbf{Novel SG Denoiser}. A graph transformer with relation-modulated attention and flow-conditioned message aggregation integrates node and edge updating.
  \item \textbf{SOTA Results.} We validate FlowSG on open-vocabulary PSG and SGG settings, and show that the mixed denoising process offers consistent gains in relation accuracy and graph-level metrics, e.g., $\sim 3$ points better than the current state-of-the-art  USG-Par~\cite{wu2025universal} across all metrics.
\end{itemize}

\section{Related Work}

\noindent\textbf{Scene Graph Generation.}
SGG methods are commonly grouped into \emph{two-stage} and \emph{one-stage} pipelines. Two-stage models first detect objects and then infer relations via message passing~\cite{he2021learning,zellers2018neural}, external knowledge~\cite{gu2021open}, or transformer-based reasoning~\cite{sortino2023transformer,he2023toward}. One-stage models~\cite{lin2020gps,cong2023reltr} formulate SGG/PSG as set prediction, jointly decoding objects and predicates in a single pass. Vision–language pretraining further improves semantic coverage~\cite{zhou2024openpsg,li2024pixels}, and panoptic SGG integrates pixel-level masks~\cite{li2024panoptic,yang2022panoptic}. Despite these advances, prevailing approaches remain \emph{deterministic and one-shot}: even with multi-round attention or message passing, decisions are finalized in a single forward pass without explicitly refining the graph structure over time. This limits error correction and weakens global structural consistency.

\noindent\textbf{Flow Matching for Graph Generation.}
Graph generative modeling has evolved from autoregressive architectures~\cite{li2023stprivacy,li2024instant3d,garg2021unconditional} and VAEs~\cite{simonovsky2018graphvae} to GANs~\cite{martinkus2022spectre}. More recently, score-based diffusion over discrete nodes/edges~\cite{kong2023autoregressive} and closely related transport formulations have achieved strong fidelity by progressively denoising combinatorial structures under graph constraints. 
Flow matching (FM) offers an alternative training objective that learns a time-dependent vector field to transport a simple prior to the data distribution~\cite{li2026efficient,li2026fixed}, with conditional variants enabling efficient guidance at inference~\cite{lipman2022flow}. Compared with diffusion process, FM avoids explicit score normalization and admits flexible ODE or SDE solvers~\cite{qin2024defog}, thereby bringing improved sampling efficiency and training stability. 
However, prior graph FM work is predominantly unconditional or weakly conditioned, and typically treats semantic labels and geometry in isolation. 
\section{Preliminaries}
\label{sec:prelim}

In this section, we briefly introduce  two techniques in flow matching: \emph{continuous flow matching} (CFM) and \emph{discrete flow} via a continuous-time Markov chain (CTMC).

\noindent\textbf{Continuous Flow Matching (CFM).}Let $x\in\mathbb{R}^{d_x}$ denote a continuous variable and $z\in\{1,\dots,K\}\cup\{\texttt{[MASK]}\}$ a categorical token (we use one-hot vectors and the probability simplex $\Delta^{K}$ for simplicity).  We couple source and target samples with a joint $\pi$ and define conditional probability paths $\{p_t\}_{t\in[0,1]}$ that interpolate between $p_0$ and $p_1$ under a monotone scheduler $\kappa_t$.
 Given a coupling $(x_0,x_1)\sim\pi$ and an interpolant $\psi_t:\mathbb{R}^{d_x}\times\mathbb{R}^{d_x}\!\to\!\mathbb{R}^{d_x}$, define the path
\begin{equation}
x_t \!\;=\!\; \psi_t(x_0,x_1), ~
u^\star(x_t,t\mid x_0,x_1) \;\!=\!\; \partial_t \psi_t(x_0,x_1).
\label{eq:cfm_target_vel}
\end{equation}
A neural vector field $v_\theta(x,t,c)$  (conditioned on image context $c$ and $t\sim \mathcal{U}[0,1]$) is trained to match the target velocity:
\begin{equation}
\mathcal{L}_{\mathrm{CFM}}\!
=\!\mathbb{E}_{(x_0,x_1)\sim\pi}
\big\|\!\,v_\theta(x_t,t,c)\!-\!u^\star(x_t,t\!\mid\! x_0,x_1)\,\big\|_2^2.
\label{eq:cfm_loss}
\end{equation}
With the common linear path $\psi_t(x_0,x_1)=(1-t)\,x_0+t\,x_1$, we have $u^\star = x_1-x_0$. At inference, we integrate the ODE
\begin{equation}
\frac{d}{dt}x_t \,=\, v_\theta(x_t,t,c),
\label{eq:cfm_ode}
\end{equation}
starting from $x_{t=0}\!\sim p_0$ (e.g., a simple Gaussian) and using a few ODE steps \cite{eijkelboom2024variational} to transport to $p_1$.

\noindent\textbf{Discrete Flow via CTMC (DFM).}
For a categorical token, let $p_t\in\Delta^{K}$ be its marginal Probability Mass Function (PMF) at time $t$. We realize a discrete-time flow with a CTMC whose time-varying \emph{rate matrix} $R_\theta(t,c)\in\mathbb{R}^{(K+1)\times(K+1)}$ satisfies $R_{ij}\!\ge\!0$ for $i\!\neq\!j$ and $R_{ii}\!=-\sum_{j\neq i}R_{ij}$~\cite{gat2024discrete}. The distribution evolves via:
\begin{equation}
\frac{d}{dt}\,p_t \;=\; p_t\,R_\theta(t,c). \nonumber
\label{eq:ctmc_master}
\end{equation}
To specify a target flow for matching, we use a two-point \emph{conditional path} between a source categorical distribution $q_0$ (masked with \texttt{[MASK]}) and the clean one-hot $q_1$:
\begin{equation}
p_t \;=\; (1-\kappa_t)\,q_0 \;+\; \kappa_t\,q_1,\qquad \kappa_t\in[0,1],\;\dot\kappa_t\ge 0.
\label{eq:disc_path} \nonumber
\end{equation}
The target instantaneous velocity is then $u^\star_t = \frac{d}{dt}p_t = \dot\kappa_t\,(q_1-q_0)$. In practice, instead of directly regressing a rate matrix to satisfy $p_tR_\theta\!\approx\!u^\star_t$, it is numerically advantageous to train a network to predict the \emph{clean posterior} $q_1=p_{1|t}(\cdot\,|\,z_t,c)$ and assemble a feasible rate matrix from it at sampling time. We therefore minimize a time-conditioned cross-entropy:
\begin{equation}
\mathcal{L}_{\mathrm{DFM}}
=\mathbb{E}_{(z_0,z_1)\sim\pi,\;t\sim\mathcal{U}[0,1]}
\Big[\,\mathrm{CE}\big(q_1,\; f_\theta(z_t,t,c)\big)\,\Big],
\label{eq:dfm_ce}
\end{equation}
where $z_t\!\sim\!p_t$ from Eq. (\ref{eq:disc_path}) and $f_\theta$ outputs logits over the $K{+}1$ categories. At inference, we either (i) integrate Eq.~(\ref{eq:ctmc_master}) with an ODE solver using a rate matrix assembled from $f_\theta$ and $\kappa_t$, or (ii) simulate jumps with an Euler scheme {\cite{qin2024defog,gat2024discrete}}; both choices leave the \emph{trained} classifier $f_\theta$ unchanged and permit schedule/solver swaps without retraining.

\noindent\textbf{Hybrid Graph-Wise Factorization.}
For a graph $G_t$ with node slots $\{x^{(n)}_t,z^{(n)}_t\}$ and edge slots $\{z^{(ij)}_t\}$, we evolve all slots in parallel while conditioning each local update on the {global} noisy graph and image features $\mathcal{C}$:
\begin{equation}
\begin{aligned}
x^{(n)}_{t} &: \;\;\frac{d}{dt}x^{(n)}_t \,=\, v_\theta^{\mathrm{node}}\!\big(x^{(n)}_t,\,t,\,G_t,\,\mathcal{C}\big),\\
p^{(n)}_t   &: \;\;\frac{d}{dt}p^{(n)}_t \,=\, p^{(n)}_t\,R_\theta^{\mathrm{node}}\!\big(t,\,G_t,\,\mathcal{C}\big),\\
p^{(ij)}_t &: \;\;\frac{d}{dt}p^{(ij)}_t \,=\, p^{(ij)}_t\,R_\theta^{\mathrm{edge}}\!\big(t,\,G_t,\,\mathcal{C}\big).
\end{aligned}
\label{eq:graph_hybrid}
\end{equation}
Conditioned independence across slots \emph{given} $(G_t,x)$ yields the standard factorization for a small step from $t$ to $\tau$:
\begin{equation}
p(G_\tau \mid G_t) \;=\; \prod_{n} p^{(n)}_{\tau\mid t}\;\;\prod_{i<j} p^{(ij)}_{\tau\mid t},
\label{eq:graph_factorization}
\end{equation}
with each factor obtained by the corresponding ODE (continuous $x$) or CTMC update (discrete $z$).

\section{Method}
\label{sec:method}
\begin{figure*}[t]
\centering
\includegraphics[width=0.95\textwidth]{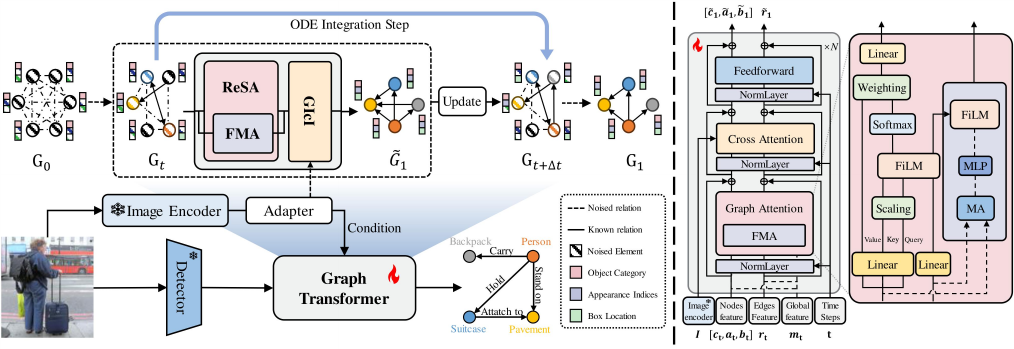}
\caption{
The overview of our \textbf{FlowSG}. \textbf{(Left)} Image-guided iterative scene graph generation via flow matching. Starting from a noised graph $G_0$, our graph transformer refines predictions through ODE integration steps ($G_t \to G_{t+\Delta t}$), outputting velocity fields for continuous bounding boxes and clean posteriors for discrete codes. \textbf{(Right)} Graph transformer architecture consisting of:  \textit{(i)} Relation-modulated Self-Attention (FiLM-based predicate injection), \textit{(ii)} Flow-conditioned Message Aggregation (ime-conditioned neighborhood message aggregation with degree-aware 
weighting), and \textit{(iii)} Global Image-condition Integration module (cross-attention to image features). 
}
\label{fig:method}
\end{figure*}

The overall architecture of \textbf{FlowSG} is depicted in Fig. \ref{fig:method}. Briefly, we construct an initial  scene graph from detected objects, discretize semantic features via a vision-language codebook, and apply hybrid discrete-continuous flow matching to iteratively refine graph structure and semantics. Sec. \S \ref{sec:quantization} details semantic discretization and Sec. \S \ref{sec:graph_flow} describes the hybrid flow matching process. \S \ref{sec:arch} describes the architecture of the denoiser.
\subsection{Scene Graph Tokenization}
\label{sec:quantization}

We cast scene-graph generation as a hybrid discrete–continuous construction process. To make the discrete part \emph{predictable} and semantics-aware, we tokenize objects and relations into compact codes; geometry remains continuous and is handled by CFM. This section details the object and relation tokenizers.

\noindent\textbf{Setup.}
Given an image $I$, a frozen detector $\Phi_{\mathrm{det}}$~\cite{cheng2022masked} produces $N$ region proposals with RoI features, class logits, and boxes:
\begin{equation}
\bigl\{(\mathbf{f}_i,\mathbf{s}_i,\mathbf{b}_i)\bigr\}_{i=1}^{N}\!
=\!\Phi_{\mathrm{det}}(I;\theta_{\mathrm{det}}),
\mathbf{f}_i\!\in\!\mathbb{R}^{d},\;\mathbf{s}_i\!\in\!\mathbb{R}^{C_{\!obj}}.
\label{eq:detector}
\end{equation}
We denote the ground-truth object class by $y_i^\star\in\{1,\dots,C_{\!obj}\}$ and the ground-truth predicate phrase for an ordered pair $(i,j)$ by $r_{ij}^\star$.

\noindent\textbf{Object Discretization.}
Operating generatively in raw feature space $\mathbb{R}^{d}$ is costly and unstable. Following discrete representation learning~\cite{lin2024instructscene}, we adopt vector quantization to discretize object visual features into compact codes in a language-aligned embedding space. Specifically, we extract region features by cropping and encoding with CLIP's frozen image encoder:
\begin{equation}
\mathbf{u}_i =   \text{CLIP}_{\text{img}}(\Phi_{\text{crop}}\bigl({I},  \mathbf{b}_i \bigr) ),
\end{equation}
where $\text{CLIP}_{\text{img}}(.)$ is the image encoder of CLIP and $\Phi_{\text{crop}}(.)$ is an image crop function given a box. 
Then we train a VQ-VAE with codebook $\mathcal{H}_{\!obj}=\{\mathbf{e}_k\}_{k=1}^{K_a}$. For each region feature $\mathbf{u}_i$, we quantize to the nearest codebook entry:
\begin{equation}
a_i^\star = \arg\min_{k\in[K_a]}\bigl\|\mathbf{u}_i-\mathbf{e}_k\bigr\|_2^2.
\label{eq:obj_quantize}
\end{equation}
The encoder, decoder, and codebook are trained with the discrete VQ-VAE objective~\cite{rolfe2016discrete,ramesh2021zero,jang2016categorical}.

Finally, geometry remains continuous and object class labels follow the dataset taxonomy. We represent the final \emph{target} node tokens  in a scene graph as
\begin{equation}
\mathbf{n}_i \;=\; \bigl(\mathbf{a}_i^\star,\; \mathbf{c}_i^\star,\; \mathbf{b}_i^\star \bigr)
~\in~ [K_a]\times [C_{\!obj}] \times \mathbb{R}^{4},
\label{eq:node_token}
\end{equation}
where $\mathbf{c}_i^\star$ is one–hot object class. During flow matching, we initialize appearance tokens as \texttt{[MASK]} and leverage category labels $y_i$ as untouched informative priors, allowing the model to progressively refine appearance codes from scratch while grounded in semantic guidance.

\noindent\textbf{Relation Discretization.}
One-hot predicate labels discard semantic proximity and hinder open-vocabulary generalization. We therefore quantize relation \emph{phrases} into a compact predicate codebook using language embeddings. Let $\mathcal{C}_{\!rel}$ be a corpus of relation words aggregated from datasets \cite{gupta2015visual,yang2022panoptic} and external corpus, e.g., ConceptNet {\cite{speer2017conceptnet}}, and let $\mathcal{T}_{\mathrm{text}}$ be CLIP's text encoder \cite{clip}:
\(
\mathbf{u}(w)\;=\;\frac{\mathcal{T}_{\mathrm{text}}(w)}{\|\mathcal{T}_{\mathrm{text}}(w)\|_2}\in\mathbb{R}^{d_t},\; w\in\mathcal{C}_{\!rel}.
\)
The quantization training loss is the same as object feature VQ-VAE. Last, for each edge $(i,j)$ with ground-truth phrase $r_{ij}^\star$, the {predicate token} is
\begin{equation}
p_{ij} \;=\; k\bigl(r_{ij}^\star\bigr)\in[K_r].
\label{eq:edge_token}
\end{equation}
Since the codebook is learned in language space, similar meaning or fine-grained phrases can map to nearby or identical codes, providing semantic smoothing and enabling open-vocabulary decoding via nearest-neighbor back to CLIP space at inference.

\subsection{Hybrid Flow Matching for SGG}
\label{sec:graph_flow}
We generate a \emph{hybrid} scene graph from a masked/noisy prior using flow matching. 
Given a prior graph $G_{0}$ with masked node/edge semantics, we refine it into a complete scene graph $G_{1}$ via the hybrid formulation in Sec.~\ref{sec:prelim}. 
Heterogeneous attributes are handled separately yet coupled through a shared, image–conditioned graph encoder: \emph{discrete semantics} $\mathbf{s}=(\mathbf{c},\mathbf{r},\textit{\textbf{a}})$ (object labels, predicate categories, appearance (visual feature) tokens) evolve with a CTMC-based discrete flow, while \emph{continuous boxes} $\mathbf{g}=\{\mathbf{b}_i\}_{i=1}^{N}$ evolve with CFM.

\noindent \textbf{Noised and Target  SG Construction.}
Formally, we write a scene graph at time $t$ as
$
G_t = (\mathcal{V}, \mathcal{E}, \mathbf{s}_t, \mathbf{g}_t), \quad
\mathbf{s}_t = (\mathbf{c}_t, \mathbf{r}_t, \textit{\textbf{a}}_t), \quad
\mathbf{g}_t = \{\mathbf{b}_i(t)\}_{i=1}^{N},
$
with node set $\mathcal{V} = \{1, \dots, N\}$ and directed edge set $\mathcal{E} \subseteq \{(i,j) \,|\, i \neq j\}$.
Here, $\mathbf{c}_t$ denotes object categories, $\mathbf{r}_t$ denotes relation types, $\mathbf{a}_t$ denotes appearance codes, and $\mathbf{g}_t$ collects bounding boxes.
Each discrete slot is modeled by a probability mass function   on a simplex that includes a \texttt{[MASK]} state:
\begin{equation}
p_t^{\mathrm{obj}}(i) \in \Delta^{K_c},~~
p_t^{\mathrm{rel}}(i,j) \in \Delta^{K_r}, ~~
p_t^{\mathrm{app}}(i) \in \Delta^{K_a},
\label{eq:slot_simplex} \nonumber
\end{equation}
where $K_c$, $K_r$, and $K_a$ denote the vocabulary sizes for objects, predicates, and appearance codes, respectively.

We initialize from a simple source $G_0 = (\mathcal{V}, \mathcal{E}, \mathbf{s}_0, \mathbf{g}_0)$ with
$
p_0^{\mathrm{obj}}(i) = \delta_{c_i^{\mathrm{det}}}, \quad
p_0^{\mathrm{rel}}(i,j) = \delta_{\texttt{[MASK]}}, \quad
p_0^{\mathrm{app}}(i) = \delta_{\texttt{[MASK]}}, \quad
\mathbf{b}_i(0) \sim \mathcal{N}(\mathbf{0}, \mathbf{I}_4),
$
where $c_i^{\mathrm{det}}$ is the object category from the pretrained detector. Notably, {object categories are not masked} but serve as priors to guide the generation of relationships and appearances,  simplifying training and enhancing stability. 
In contrast, relation types and appearance codes are fully masked, and bounding boxes are initialized from standard Gaussian noise. 
Similarly, the target scene graph 
\(G_1 = (\mathcal{V}, \mathcal{E}, \mathbf{s}_1, \mathbf{g}_1)\) is defined with 
\begin{equation}
    \begin{aligned}
        p_1^{\mathrm{obj}}(i) = \delta_{y_i^\star}, \quad
p_1^{\mathrm{rel}}(i,j) = \delta_{r_{ij}^\star}, \\
p_1^{\mathrm{app}}(i) = \delta_{a_i^\star}, \quad
\mathbf{b}_i(1) = \mathbf{b}_i^\star,
    \end{aligned}
\end{equation}
where $y_i^\star \in [K_c]$ are ground-truth object classes, $r_{ij}^\star \in [K_r]$ are ground-truth predicate types, $a_i^\star \in [K_a]$ are CLIP-aligned appearance codes from the semantic codebook $\mathcal{H}_{\text{obj}}$ (Sec.~\ref{sec:quantization}), and $\mathbf{b}_i^\star \in \mathbb{R}^4$ are ground-truth bounding boxes.


\noindent \textbf{Hybrid Flow.}
The hybrid flow transports $(\mathbf{s}_0, \mathbf{g}_0) \mapsto (\mathbf{s}_1, \mathbf{g}_1)$ by integrating the continuous flow matching (CFM) ODE for geometric attributes $\mathbf{g}$ (Eqs.~\eqref{eq:cfm_path_geo}–\eqref{eq:cfm_vel_geo}, \eqref{eq:cfm_ode}) and evolving the categorical marginals for semantic attributes $\mathbf{s}$ under the continuous-time Markov chain (Eq.~\eqref{eq:ctmc_master}). Both processes are conditioned on visual context $\mathcal{C}$ via a  graph transformer. The global visual context $\mathcal{C}$ is encoded by a frozen CLIP visual encoder with an adapter (see Figure \ref{fig:method}).
 
\noindent \textbf{Flow Path Construction}
We interpolate between a simple \emph{source} at $t{=}0$ and the \emph{target} at $t{=}1$ using a monotone scheduler $\kappa_t\!\in\![0,1]$ (e.g., $\kappa_t=1-\cos(\tfrac{\pi t}{2})$).

\emph{CFM for continuous boxes.}
Let $\mathbf{g}_1$ be the clean box configuration and draw $\mathbf{g}_0\!\sim\!\mathcal{N}(\mathbf{0},\mathbf{I})$. Using the conditional interpolant of Sec.~\ref{sec:prelim},
\begin{equation}
\mathbf{g}_t \;=\; \psi_t(\mathbf{g}_0,\mathbf{g}_1)\;=\; (1-\kappa_t)\,\mathbf{g}_0 + \kappa_t\,\mathbf{g}_1,
\label{eq:cfm_path_geo}
\end{equation}
the target velocity is
\begin{equation}
\mathbf{u}^{\star}_{\mathbf{g}}(\mathbf{g}_t,t)\;=\;\partial_t \psi_t\;=\;\dot\kappa_t\,(\mathbf{g}_1-\mathbf{g}_0).
\label{eq:cfm_vel_geo}
\end{equation}

\emph{DFM for discrete semantics.}
Let $\mathbf{s}_0$ denote the masked prior (all \texttt{[MASK]}) and $\mathbf{s}_1$ the clean assignment. As in Eq.~\eqref{eq:disc_path}, we use the two-point conditional path
\begin{equation}
p_t(\mathbf{s}\mid \mathbf{s}_0,\mathbf{s}_1) \;=\; (1-\kappa_t)\,\delta_{\mathbf{s}_0}(\mathbf{s}) \;+\; \kappa_t\,\delta_{\mathbf{s}_1}(\mathbf{s}),
\label{eq:dfm_path_sgg}
\end{equation}
whose \! slot-wise marginals evolve under the CTMC in \eqref{eq:ctmc_master}.

\noindent \textbf{Flow Velocity Prediction.}  A frozen image encoder (e.g., CLIP's image encoder \cite{clip}) first extracts global visual condition $\mathcal{C}$ from $I$. Then, we use a graph transformer $\mathcal{F}_{gtr}$  to encode the current state $(G_t,I)$, from which we predict (i) a geometric vector field for CFM and (ii) time-conditioned clean posteriors for DFM.

\noindent\emph{Geometry head.}
We predict $\mathbf{v}_\theta(\mathbf{g}_t,t,\mathcal{C})$ and train it to match Eq. \eqref{eq:cfm_vel_geo} using the CFM loss in Eq.~\eqref{eq:cfm_loss}; at sampling, we integrate the ODE in Eq.~\eqref{eq:cfm_ode} starting from $\mathbf{g}_{t=0}$.

\noindent\emph{Semantic head.}
For each node $i$ and edge $(i,j)$, $\mathcal{F}_{gtr}$  produces time-conditioned clean posteriors
$p_{1|t}\!\big(c_i \mid G_t,\mathcal{C}\big),\quad
p_{1|t}\!\big(a_i \mid G_t,\mathcal{C}\big),\quad
p_{1|t}\!\big(r_{ij} \mid G_t,\mathcal{C}\big),
$
which factorize across slots as in Eq.~\eqref{eq:graph_factorization}. We assemble a feasible, time-varying rate matrix $R_\theta(t,\mathcal{C})$ from these posteriors and $\kappa_t$ to evolve the categorical marginals via Eq.~\eqref{eq:ctmc_master}. Training follows the time-conditioned cross-entropy of Eq.~\eqref{eq:dfm_ce} with targets given by the clean tokens.



\subsection{Denoiser Architecture} \label{sec:arch}
We parameterize the hybrid flow with a graph-aware DiT-style \cite{peebles2023scalable} Transformer \(f_\theta\) (Fig.~\ref{fig:method}). 
At time \(t\!\in\![0,1]\), node and edge embeddings are initialized as
\begin{equation}
\mathbf{h}^{(0)}_i
=\big[\mathrm{Emb}(c_i^t)\ \oplus\ \mathrm{Emb}(a_i^t)\ \oplus\ \mathrm{Enc}(\mathbf{b}_i^t) \big], \nonumber
\label{eq:init_embeddings}
\end{equation}
 and \(\mathbf{e}^{(0)}_{ij}=\mathrm{Emb}(r_{ij}^t)\), where $\mathrm{Emb(.)}$ is an embedding function (e.g., the codebook of VQ-VAE) to convert the discrete code to a continuous vector. A sinusoidal time code \(\phi(t)\) modulates all blocks via adaptive layer normalization (AdaLN) \cite{peebles2023scalable}.
We stack \(L\) transformer blocks; each block contains:
(i) \emph{global image–condition integration} via cross-attention, {following image conditional guidance~\cite{yang2024open,xu2023open}}; (ii) \emph{relation-modulated self-attention}, which injects predicate semantics into node updates; and (iii) a \emph{flow-conditioned message aggregation} layer that explicitly couples nodes and edges with time-aware graph statistics. We next detail modules (ii) and (iii).

\noindent\textbf{Relation-modulated Self-attention (ReSA).}
Let \(\mathbf{h}_i^{(\ell)}\) and \(\mathbf{e}_{ij}^{(\ell)}\) denote node and edge states at block \(\ell\).
With queries/keys \(\mathbf{q}_i=W_q \mathbf{h}_i^{(\ell)}\), \(\mathbf{k}_j=W_k \mathbf{h}_j^{(\ell)}\),
we compute attention using an edge-conditioned FiLM bias \cite{perez2018film}:
\begin{equation}
\begin{aligned}
\alpha_{ij}(t)=\mathrm{softmax}_j\!\left(
\frac{\mathbf{q}_i^\top \mathbf{k}_j}{\sqrt{d}}
+ \mathrm{FiLM}\!\big(\mathbf{e}_{ij}^{(\ell)}\big)
\right),
\label{eq:rel_attn}
\end{aligned}
\end{equation}
Node values \(\mathbf{v}_j=W_v\mathbf{h}_j^{(\ell)}\) are aggregated with \(\alpha_{ij}(t)\) and passed to the feed-forward sublayer.

\noindent \textbf{Flow-conditioned Message Aggregation (FMA).}
Intuitively, aggregation should match the denoising stage: early \(t\!\approx\!1\) requires conservative, robust summaries; late \(t\!\to\!0\) benefits from sharper, higher-order cues \cite{Peebles2023DiT,Karras2022EDM}.
We therefore condition the neighborhood summary on time, degree, and local graph context. 
Let the neighbor set be \(\mathcal{N}(i)\), with degree \(\deg(i,t)\), and define the average relation context
\begin{equation}
\begin{aligned}
    \bar{\mathbf{r}}_i^{\ell}(t)=\frac{1}{|\mathcal{N}(i)^{\ell}|}\sum_{j\in\mathcal{N}(i)^{\ell}} \mathbf{e}_{ij}(t)^{\ell},
\\
\zeta_i(t)^{\ell}=\big[\ \phi(t)\ \oplus\ \log(1+\deg(i,t)^{\ell}) \oplus\ \bar{\mathbf{r}}_i^{\ell}(t)\ \big], \nonumber
\end{aligned}
\label{eq:zeta}
\end{equation}
where $\mathrm{deg(i,t)}$ is the degree of node $i$. 
We maintain a bank of permutation-invariant moment operators \(\{\Psi_k\}\) (e.g., mean, variance, skewness, kurtosis.
The time-conditioned neighborhood message aggregation is
\begin{equation}
M_i(t)^{\ell}=\sum_{k}\mathrm{softmax}\big(W_\beta\,\zeta_i(t)^{\ell}\big)\,\cdot \Psi_k\big(\{\mathbf{v}_j\}_{j\in\mathcal{N}(i)}\big) \nonumber
\label{eq:Mi}
\end{equation}
which favors robust statistics over PNA \cite{corso2020principal} when \(\zeta_i(t)^{\ell}\) indicates high noise/degree and shifts to sharper moments as the graph cleans.
We produce the node update via 
\begin{equation}
\mathbf{h}_i^{(\ell+1)}=\mathbf{h}_i^{(\ell)}+\mathrm{MLP}\!\Big(\big[\ \mathrm{AdaLN}_{\phi(t)}(M_i(t)) ^{\ell}\big]\Big). \nonumber
\label{eq:node_update}
\end{equation}

To couple node denoising and relation inference, we refine an edge hidden state \(\mathbf{e}_{ij}^{(\ell)}\) using both endpoint features and their time-conditioned summaries:
\begin{equation}
\begin{aligned}
    \mathbf{z}_{ij}(t)=\big[ \mathbf{h}_i^{(\ell+1)} \oplus \mathbf{h}_j^{(\ell+1)} \oplus\ M_i(t)^{\ell} \oplus M_j(t)^{\ell} \oplus \phi(t) \big], \\
\mathbf{e}_{ij}^{(\ell+1)}=\mathbf{e}_{ij}^{(\ell)}+\mathrm{MLP}_e\!\big(\mathrm{AdaLN}_{\phi(t)}(\mathbf{z}_{ij}(t))\big).
\label{eq:edge_state}
\nonumber
\end{aligned}
\end{equation}
Then, edge logits combine  FiLM bias and the refined edge:
\begin{equation}
\mathbf{e}_{ij}^{(\ell+1)}(t)= \mathrm{FiLM}\!\big(\mathrm{Emb}(\mathbf{e}_{ij}^{(\ell+1)})\big)\ +\ \mathbf{W}_e\,\mathbf{e}_{ij}^{(\ell+1)}.
\label{eq:edge_logits}
\end{equation}

\noindent \textbf{Training Objective.}
We follow the flow–matching formulation from Sec.~\ref{sec:prelim}: match target velocities for continuous variables and learn time–conditioned clean posteriors for discrete variables.

\emph{Continuous geometry (CFM).}
With the conditional path $\mathbf{g}_t=\psi_t(\mathbf{g}_0,\mathbf{g}_1)=(1-\kappa_t)\mathbf{g}_0+\kappa_t\mathbf{g}_1$ and target velocity $\mathbf{u}^\star_{\mathbf{g}}=\partial_t\psi_t=\dot\kappa_t(\mathbf{g}_1-\mathbf{g}_0)$, we minimize
\begin{equation}
\mathcal{L}_{\mathrm{CFM}}
=\mathbb{E}_{t\sim\mathcal{U}[0,1]}
\Big\|\,\mathbf{v}_\theta(\mathbf{g}_t,t,c)-\dot\kappa_t(\mathbf{g}_1-\mathbf{g}_0)\,\Big\|_2^2,
\label{eq:loss_cfm_consistent}
\end{equation}
where $c$ is the shared context from $(G_t,I)$.

\emph{Discrete semantics (DFM).}
Let $\mathbf{s}_0$ be the masked prior and $\mathbf{s}_1$ the clean assignment. Sampling $\mathbf{s}_t\!\sim\!p_t(\mathbf{s}\mid\mathbf{s}_0,\mathbf{s}_1)$ from Eq.~\eqref{eq:dfm_path_sgg}, a classifier $f_\theta$ predicts clean posteriors $p_{1|t}$ for each slot. We train with time–conditioned cross–entropy (factorized over nodes and edges in Eq.~\eqref{eq:graph_factorization}):
\begin{equation}
\begin{split}
\mathcal{L}_{\mathrm{DFM}} \!=
- \!\!&\sum_i \sum_{m=1}^{n_o} \log p_{1|t}(a_{i,m}^1 \mid G_t, \mathcal{C}) \\
- \!\!&\sum_{(i,j)} \sum_{m=1}^{n_r} \log p_{1|t}(r_{ij,m}^1 \mid G_t, \mathcal{C}) \Big].
\end{split}
\label{eq:loss_dfm_consistent}
\end{equation}
The overall objective is 
$\mathcal{L} \;=\; \mathcal{L}_{\mathrm{CFM}} \;+\; \lambda\,\mathcal{L}_{\mathrm{DFM}}.
\label{eq:total_loss_consistent}
$

\section{Experiments}
\subsection{Experimental Setup and Datasets}

\noindent\textbf{Datasets.} We conduct experiments on Visual Genome (VG)~\cite{krishna2017visual} and Panoptic Scene Graph (PSG)~\cite{yang2022panoptic}. \textbf{VG} is the standard scene graph benchmark with $108{,}077$ images. We use the VG150 split ($150$ object categories, $50$ predicates), comprising $\sim$75K training and $\sim$32K test images. \textbf{PSG} extends to panoptic segmentation with pixel-level masks. Derived from COCO, it contains $48{,}749$ images ($46{,}563$ train, $2{,}186$ test) across $80$ thing categories, $53$ stuff categories, and $56$ predicates.

\noindent\textbf{Evaluation Protocol.} We benchmark two standard SGG subtasks: \emph{PredCls} (ground-truth boxes and labels provided; predict predicates between object pairs) and \emph{SGDet} (jointly detect objects, classify them, and infer pairwise relations). Performance is reported using Recall (R$@$K) and mean Recall (mR$@$K) on the VG and PSG test sets.
In the closed-set setting, all relation categories are seen during training. For the open-vocabulary (OV) setting, we follow prior work \cite{he2022towards,zhou2024openpsg,chen2023expanding}: relation categories are split into base and novel at a 7:3 ratio; models are trained only on base relations and evaluated for generalization to novel ones.

\noindent\textbf{Implementation Details.}
We instantiate our model with five Transformer blocks, each equipped with $8$-head self-attention, a hidden dimension of $512$, and a dropout rate of $0.1$. For visual conditioning, we adopt a frozen CLIP ViT-B/$16$ image encoder~\cite{clip} to extract image features for cross-attention, and follow Detector~\cite{cheng2022masked} by using predictions from a frozen detector as object priors to ensure fair comparison. Both the object and relation codebooks contain $64$ entries with dimensionality $512$. We use $4$ ordered indices to quantize object appearance features and $4$ ordered indices to encode relation categories, following~\cite{lin2024instructscene}.
To improve training robustness, we introduce a stochastic edge-only refinement mode: with probability $0.2$, we keep node attributes fixed and only generate edges (relations). The model is optimized with AdamW~\cite{loshchilov2017decoupled} for $500$K iterations using a batch size of $128$, a learning rate of $1\times 10^{-4}$, and a weight decay of $0.02$. All experiments are run on $4$ NVIDIA A$100$ GPUs.

\begin{table}[t]
\centering
\resizebox{\columnwidth}{!}{
\begin{tabular}{l|cc|cc}
\toprule
 \multicolumn{1}{c|}{\multirow{2}{*}{Methods}} & \multicolumn{2}{c|}{SGDet} & \multicolumn{2}{c}{PredCls} \\ 
\cmidrule{2-5} 
 & R/mR@50 & R/mR@100 & R/mR@50 & R/mR@100 \\ 
\midrule
\rowcolor{gray!15}\multicolumn{5}{l}{\textbf{\textit{PSG Dataset}}} \\

 IMP\cite{imp} & 18.2 / \,7.1 & 18.6 / \,7.2 & 36.8 / 10.9 & 38.9 / 11.6 \\ 
 MOTIF\cite{zellers2018neural} & 21.7 / \,9.6 & 22.0 / \,9.7 & 50.4 / 22.1 & 52.4 / 22.9 \\
 VCTree\cite{tang2019learning} & 22.1 / 10.2 & 22.5 / 10.2 & 50.8 / 22.6 & 52.7 / 23.3 \\
 ADtrans\cite{li2024panoptic} & 29.6 / 29.7 & 30.0 / 30.0 & \,\,\,—\,\, / 36.2 & \,\,\,—\,\, / 38.8 \\
 PairNet\cite{wang2024pair} & 35.6 / 28.5 & 39.6 / 30.6 & \,\,\,—\,\, / \,\,—\,\,\, & \,\,\,—\,\, / \,\,—\,\,\, \\
 SPADE$^\dag$\cite{hu2025spade} & {43.8} / {38.9} & {49.3} / \underline{46.5} & {64.2} / {49.1} & {69.3} / {54.8} \\
 USG-Par$^\dag$\cite{wu2025universal} & \underline{44.6} / \underline{40.9} & \underline{51.3} / {42.7} & \underline{67.2} / \underline{51.1} & \underline{72.3} / \underline{57.8}\\
 
 \midrule
\rowcolor{blue!15}
 \textbf{FlowSG } & \textbf{46.3} / \textbf{42.7} & \textbf{53.3} / \textbf{48.3} & \textbf{69.4} / \textbf{54.9} & \textbf{74.3} / \textbf{61.3} \\
\midrule
\rowcolor{gray!15}\multicolumn{5}{l}{\textbf{\textit{Visual Genome Dataset}}} \\

 MOTIF\cite{zellers2018neural} & 32.5 / \,6.6 & 36.8 / \,7.9 & 65.3 / 14.9 & 67.2 / 16.3 \\
 VCTree\cite{tang2019learning} & 31.9 / \,6.4 & 36.0 / \,7.3 & \underline{65.5} / 16.7 & {67.4} / 17.9 \\ 
 PE-Net\cite{Zheng_2023_CVPR} & 26.5 / \underline{16.7} & 30.9 / \underline{18.8} & 59.0 / 38.8 & 61.4 / 40.7 \\
 OpenPSG$^\dag$\cite{zhou2024openpsg} & {32.7} / {13.5} & 38.0 / {18.3} & 62.2 / \underline{45.8} & 67.4 / 50.3 \\
 DSGG\cite{hayder2024dsgg} & \underline{32.9} / 13.0 & \underline{38.5} / 17.3 & 53.9 / {39.4} & 65.1 / \underline{49.9} \\
 CAPSGG\cite{huang2025navigating} & 18.3 /  \,\,—\,\,\, & 22.8 /  \,\,—\,\,\, & {63.4} /  \,\,—\,\,\, & \underline{67.7} /  \,\,—\,\,\, \\
 \midrule
\rowcolor{blue!15}
 \textbf{FlowSG } & \textbf{36.5} / \textbf{18.4} & \textbf{42.4} / \textbf{21.6} & \textbf{65.7} / \textbf{49.2} & \textbf{68.8} / \textbf{53.3} \\
\bottomrule
\end{tabular}
}
\caption{Results of two-stage methods on PSG and VG under the closed-set protocol. $\dagger$ denotes reproduced numbers. Best and second-best are marked in \textbf{bold} and \underline{underlined}, respectively.
}
\label{tab:two_stage_results}
\end{table}

\begin{table}[t]
\centering
\resizebox{0.93\columnwidth}{!}
{
\begin{tabular}{l|cc|cc}
\toprule
Methods & R@50 & mR@50 & R@100 & mR@100 \\ 
\midrule
SGTR~\cite{li2022sgtr} & 24.6 & 12.0 & 28.4 & 15.2 \\
EGTR~\cite{im2024egtr} & 30.2 & 5.5 & 34.3 & 7.9 \\
SpeaQ~\cite{kim2024groupwise} & 32.9 & 11.8 & 36.0 & 14.1 \\
OvSGTR$^\dag$~\cite{chen2023expanding} & 33.8 & 7.2 & 37.3 & 8.8 \\
Hydra-SGG~\cite{chen2025hydrasgg} & 28.6 & 15.9 & 33.4 & 19.4 \\
HRTrans~\cite{fu2025hybrid} & \underline{34.1} & \underline{16.0} & \underline{38.3} & \underline{20.5} \\
\midrule
\rowcolor{blue!15}
\textbf{FlowSG} & \textbf{36.5} & \textbf{18.4} & \textbf{42.4} & \textbf{21.6} \\
\bottomrule
\end{tabular}
}
\caption{Performance of one-stage methods on VG dataset on SGDet task in the closed-set scenario.}
\label{tab:one_stage_results}
\vspace{-5mm}
\end{table}

 \subsection{Main Results}
\noindent\textbf{VG Results.}
We evaluate \textbf{FlowSG} on Visual Genome (VG) under both closed- and open-set protocols using the same pre-trained detector for \emph{SGDet} (Tabs.~\ref{tab:two_stage_results}, \ref{tab:one_stage_results}, \ref{T:ovdr}). In the \emph{closed-set} setting, \textbf{FlowSG} sets a new state of the art among {two-stage} systems on SGDet with \(\sim\!3\)--\(4\) point gains in \(\text{R@50/100}\), while remaining competitive on \(\text{mR@50/100}\). On \emph{PredCls} it achieves the best \(\text{mR@50/100}\) and recall, indicating improved long-tail robustness without sacrificing head performance. Against strong \emph{one-stage} models, FlowSG attains the best \(\text{R@50/100}\) and \(\text{mR@50/100}\) with about \(2\) point improvements. In the \emph{open-set} evaluation (Tab.~\ref{T:ovdr}), FlowSG further strengthens generalization to unseen predicates, surpassing VL-IRM on \(\text{mR@50/100}\) and lifting \(\text{R@100}\) by about \(+2\), validating the benefit of progressive, image-conditioned generation.

\noindent\textbf{PSG Results.}
On PSG (Tabs.~\ref{tab:two_stage_results}, \ref{T:ovdr}), \textbf{FlowSG} leads across protocols. In the \emph{closed-set two-stage} setting, it improves SGDet \(\text{R/mR@50/100}\) by roughly \(+1\)--\(2\) points over prior strong baselines (e.g., USG-Par\cite{wu2025universal}), and on \emph{PredCls} widens the gap further with about \(\!+2\) in recall and \(\sim\!2\)--\(3\) in mean-recall, showing better coverage of rare relations while preserving head accuracy. In the \emph{open-set} scenario, FlowSG attains the highest \(\text{mR@50/100}\), surpassing VL-IRM by about \(4\) and \(2\) points, respectively. Overall, FlowSG consistently improves recall and tail robustness on both VG and PSG across evaluation metrics.

\begin{table}[t]
\centering
\resizebox{\columnwidth}{!}
{

\begin{tabular}{l|l|cccc}
\toprule
{Datasets} & {Methods} &  R@50 &mR@50 &R@100 & mR@100 \\ \midrule
\multirow{5}{*}{PSG} & PGSG\cite{li2024pixels} & 15.5 & 10.1 & 17.7 & 11.5 \\
 & OvSGTR$^*$\cite{chen2023expanding} & 19.3 & 12.4 & 22.8 & 14.0 \\
 & OpenPSG\cite{zhou2024openpsg} & {21.2} & \underline{19.8} & {25.1} & {21.4} \\
& VL-IRM$^*$\cite{min2025vision} & \underline{25.1} & 18.2 & \underline{29.3} & \underline{22.5} \\
 \cmidrule{2-6}
 \rowcolor{blue!15}
\cellcolor{white} & \textbf{FlowSG} & \textbf{26.7} & \textbf{22.3} & \textbf{31.8} & \textbf{24.2} \\ \midrule
 
\multirow{6}{*}{VG} 
 & VS\(^3\)\cite{zhang2023learning} & 15.6 & 6.7 & {17.2} & 7.4 \\
 & PGSG\cite{li2024pixels} & \underline{15.8} & 5.2 & 19.1 & 7.3 \\
 & OvSGTR$^*$\cite{chen2023expanding} & {15.1} & 5.3 & 19.3 & 7.5 \\ 
 & VL-IRM$^*$\cite{min2025vision} & 14.1 & \underline{8.4} & \underline{20.4} & \underline{12.7} \\

 \cmidrule{2-6} 
\rowcolor{blue!15}
 \cellcolor{white} & \textbf{FlowSG} & \textbf{16.9}& \textbf{9.7} & \textbf{22.5} &  \textbf{14.1} \\ 
\bottomrule
\end{tabular}}
\caption{Compared to the state-of-the-art PSG and SGG models on the  VG and PSG  dataset in the  open-set \cite{chen2023expanding} scenario. $^*$ denotes training with the same dataset.  }
\label{T:ovdr}
\vspace{-5mm}
\end{table}

\subsection{Ablation Study}
\label{Ablation Study}

 In this section, we study three components: the setting of the denoiser, sampling strategy, and tokenization design.

\noindent\textbf{Graph Transformer Components.} We ablate the proposed graph transformer on PSG under SGDet (Tab.~\ref{tab:ablation_gt}). Removing the \emph{entire} Flow-conditioned Message Aggregation (FMA) block 
causes a marked drop across all metrics, confirming that flow-aware message passing is central to our gains. The fine-grained variants show that \emph{both} edge- and node-level aggregations are necessary: eliminating either yields consistent declines ($3-6$ points), indicating complementary roles in refining relation cues and object context. 
Finally, discarding cross-attention to global visual features leads to the \emph{largest} degradation $7-11$ points across R/mR), underscoring the importance of image guidance for disambiguating relations. Together, these studies validate our design choices and highlight the synergy between flow-conditioned aggregation and visual guidance.

\noindent\textbf{Sampling Strategies.} 
We compare four initial distributions for the discrete states at $t\!=\!0$: \textit{Uniform}~\cite{campbell2024generative} ($p_0 = [\frac{1}{Z}, \ldots, \frac{1}{Z}]$): probability uniformly distributed across all states. \textit{Masking}~\cite{campbell2024generative} ($p_0 = [0, \ldots, 0, 1]$): all mass in a special mask state. \textit{Marginal}~\cite{vignac2023digress,lin2024instructscene,xu2024discrete} ($p_0 = [m_1, \ldots, m_Z]$): $m_i$ is the dataset marginal probability of state $i$. \textit{Absorbing} ($p_0 = [0, \ldots, 1, \ldots, 0]$): one-hot encoding of the most common state.

From the results, \textbf{Marginal} initialization matching dataset priors consistently attains the best R/mR@$50$/$100$, with clear gains over the next-best alternatives. \textbf{Absorbing} yields competitive recall but noticeably lower mean-recall, indicating a head-class bias detrimental to tail relations. \textbf{Uniform} and \textbf{masking} underperform overall, likely because they either discard corpus statistics or introduce unnecessary mask-token complexity. The improvements are more pronounced on mR than R, underscoring that principled initialization is key for long-tail predicate recognition. We therefore adopt \emph{marginal} for categorical tokens and Gaussian noise for continuous boxes.

\begin{table}[t]
\centering
\caption{Ablation of graph transformer components on PSG under the closed-set SGDet; “MA” denotes message aggregation.}
\resizebox{\columnwidth}{!}{
\begin{tabular}{l|cc|cc}
\toprule
 Variant & R@50 & mR@50 & R@100 & mR@100 \\

\midrule
\rowcolor{gray!15}\multicolumn{5}{l}{\textit{FMA Module Analysis}} \\
$\mathrm{w/o ~FMA}$ &  40.5 & 37.1 & 47.2 & 39.7\\ 
$\mathrm{w/o ~Edge MA}$ & 43.1 & 38.5 & 49.8 & 44.2 \\
$\mathrm{w/o~ Node MA}$ & 42.8 & 38.9 & 49.6 & 42.5 \\
\midrule
\rowcolor{gray!15}\multicolumn{5}{l}{\textit{Visual Guidance}} \\
$\mathrm{w/o ~Cross}$-$\mathrm{attn}$ & 39.2 & 34.3 & 45.3 & 36.9 \\
\midrule
\textbf{FlowSG} & \textbf{46.3} & \textbf{42.7} & \textbf{53.3} & \textbf{48.3} \\
\bottomrule
\end{tabular}
}
\label{tab:ablation_gt}
\end{table}

\begin{figure}[t]
  \includegraphics[width=.45\textwidth]{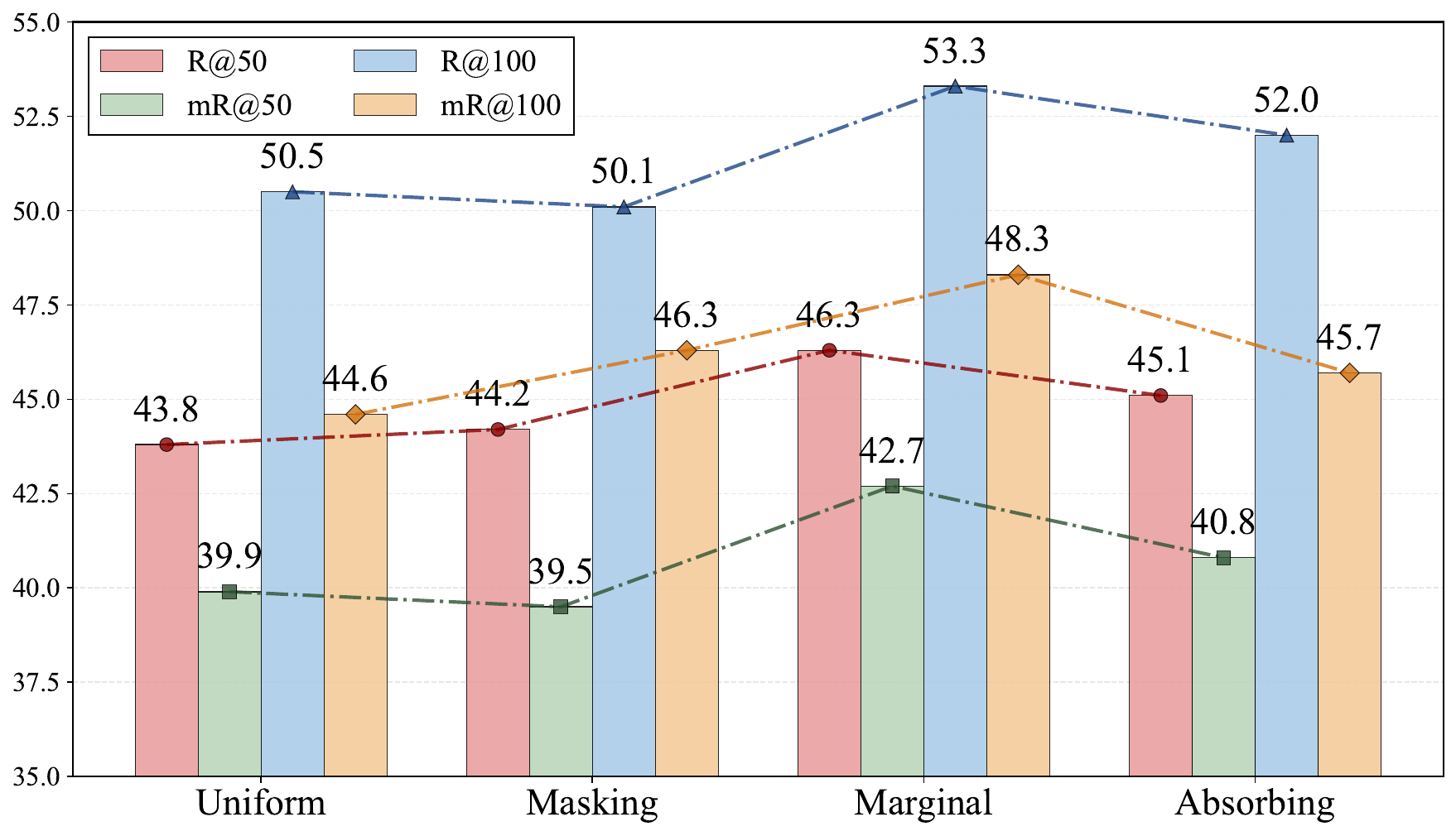}
  \centering
  \caption{Results of four sampling strategies on  closed-set PSG. 
  }
  \label{fig:sample}
\end{figure}


\noindent\textbf{Tokenization Design.}
On PSG (Tab.~\ref{tab:tokenization}), we ablate appearance codebook capacity and slot factorization. Increasing the codebook from \(\mathbf{32{\times}256}\) to \(\mathbf{64{\times}256}\) yields double-digit gains in both recall and mean–recall, indicating that a small codebook is an information bottleneck. Further enlarging the embedding to \(\mathbf{64{\times}512}\) brings consistent improvements, suggesting richer codewords better capture fine-grained appearance. For structure, using \(\mathbf{M{=}4}\) slots outperforms \(\mathbf{M{=}3/5}\). 
A limited number of slots may reduce the ability to express object features, while an excessive number significantly increases the difficulty of description.
We therefore adopt the \(\mathbf{64{\times}512}\) codebook with \(\mathbf{M{=}4}\) ordered slots as our default.

\subsection{Visualization Results}
\begin{figure}[t]

  \includegraphics[width=.45\textwidth]{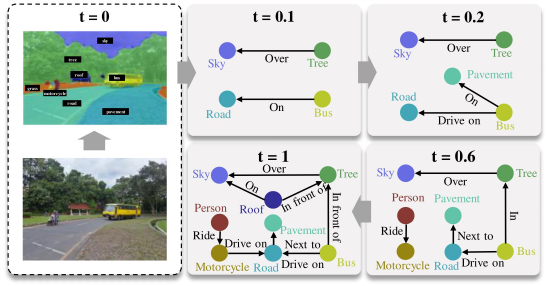}
  \centering
  \caption{Progressive \textbf{FlowSG}. From the input image (bottom-left) and ground truth at \(t{=}0\) (top-left), we show predictions refined at \(t\in{0.1,0.2,0.4,0.6}\).}
  \label{fig:vis}
\end{figure}

Figure~\ref{fig:vis} visualizes FlowSG’s image-conditioned transport over time. Early steps (\(t{=}0.1,0.2\)) establish nodes and coarse relations but include generic or misplaced edges. Mid steps refine structure, repairing relation endpoints and correcting predicates (e.g., “On” \(\rightarrow\) “Drive on”). The final step produces a compact, semantically consistent graph close to ground truth. This coarse-to-fine trajectory shows that flow matching performs targeted, structure-aware edits, outperforming one-shot prediction in stability.


\begin{table}[t]
\centering
\caption{Ablation of tokenization on PSG: \textit{(top)} effect of codebook size \(K{\times}d\); \textit{(bottom)} slot factorization levels \(M\).
}
\resizebox{0.92\columnwidth}{!}
{
\begin{tabular}{l|cc|cc}
\toprule
Configuration & R@50 & mR@50 & R@100 & mR@100 \\
\midrule
\rowcolor{gray!15}\multicolumn{5}{l}{\textit{Size (K) and Dimension (d)}} \\
32$\times$256  & 32.7 & 28.5 & 39.5 & 31.2 \\
64$\times$256  & 43.3 & 41.1 & 52.1 & 46.6 \\
\textbf{64$\times$512 (Ours)} & \textbf{46.3} & \textbf{42.7} & \textbf{53.3} & \textbf{48.3} \\
\midrule
\rowcolor{gray!15}\multicolumn{5}{l}{\textit{ Slots (M)}} \\
M=3  & 44.1 & 38.8 & 50.8 & 45.5 \\
M=5 & 35.6 & 31.4 & 42.3 & 47.0 \\
\textbf{M=4 (Ours)} & \textbf{46.3} & \textbf{42.7} & \textbf{53.3} & \textbf{48.3} \\
\bottomrule
\end{tabular}
}
\label{tab:tokenization}
\end{table}

\subsection{Conclusion}

In this work, we cast SGG as progressive transport on a {hybrid} state—discrete tokens (objects/predicates) and continuous geometry (boxes). FlowSG couples (i) {hybrid flow matching}, where a probability-flow ODE drives boxes while a discrete-flow objective updates categorical tokens, with (ii) a {graph-aware   transformer} equipped with {Flow-conditioned Message Aggregation (FMA)}. FMA performs relation-modulated self-attention with FiLM gating, then aggregates {edge- and node-level} messages, and finally injects {visual guidance} via cross-attention to image features, enabling semantics–geometry co-generation. Extensive results demonstrate FlowSG superior performance on a wide-range of datasets.
\textbf{Limitations.} The few-step ODE integration still adds inference cost compared to strictly one-shot models.  In  future work, We plan to (i) train end-to-end with detector heads, (ii) explore model compression, step-adaptive solvers, and early-exit policies for the denoiser.

\section*{Acknowledgments}
This research was partially supported by the National
Natural Science Foundation of China (NSFC) (62306064), the Sichuan Science and Technology Program (2023ZYD0165,
2024ZDZX0011 and 2024ZHCG0009) and the Tianfu Jiangxi Laboratory(TFJX-ZD-2024-001). We appreciate all the authors for their
fruitful discussions. 
{
    \small
    \bibliographystyle{ieeenat_fullname}
    \bibliography{main}
}


\end{document}